\documentclass[runningheads]{llncs}
\usepackage[T1]{fontenc}

\usepackage{marvosym}
\usepackage{orcidlink}
\usepackage{graphicx}
\usepackage{amsmath}
\usepackage{amssymb}
\usepackage{booktabs}
\usepackage{multirow}
\usepackage{subcaption}
\begin{document}
\title{Automatic Echocardiography Segmentation via Transition Probability Correlation for Stable Semantic Extraction}
\titlerunning{Echocardiography Segmentation via Transition Probability Correlation}

\author{
Xinran Chen\textsuperscript{\textdagger}\orcidlink{0009-0009-5111-3260}
\and
Xiyuan Wang\textsuperscript{\textdagger}\orcidlink{0000-0002-3677-3596}
\and
Guangquan Zhou\textsuperscript{\Letter} \orcidlink{0000-0002-6467-3592}
\and
Chuan Chen\textsuperscript{\Letter} \orcidlink{0000-0001-5701-0857}
}

\authorrunning{X. Chen et al.} 

\institute{
School of Biological Science and Medical Engineering, Southeast University, Nanjing, China\\
\email{guangquan.zhou@seu.edu.cn, chuanchen@seu.edu.cn}
}

\maketitle

\begingroup
\renewcommand\thefootnote{}
\footnotetext{
\textsuperscript{\textdagger} These authors contributed equally.
}
\endgroup

\begin{abstract}
While echocardiography is essential for cardiovascular diagnosis, inherent speckle noise and low signal-to-noise ratio often lead to ambiguous semantic features and fragmented boundaries. These limitations significantly hinder the segmentation accuracy of deep learning models in complex clinical cases. Moreover, temporal motion of the heart plays a critical role in recognizing anatomical structures. To address these challenges, we designed a STLSF module which comprises a window-matching-based semantic correction component and a semantics-guided texture enhancement component. By leveraging local transition probability correlations to correct semantics and employing semantics-guided texture enhancement, the STLSF module effectively mitigates texture instability and ambiguous semantic interpretations caused by disadvantaged echocardiography quality. Additionally, to facilitate the encoder's adaptation to the intrinsic priors of ultrasound-specific imaging patterns, we propose a frequency-aware denoising pre-training method. The entire work builds a convolution-based network with locality inductive bias and long-range dependencies. Extensive experiments confirm our SOTA performance, achieving 93.87\% Dice on CAMUS and 92.62\% on EchoNet-Dynamic, with respective HD95 values of 3.29mm and 2.73mm.
\end{abstract}

\section{Introduction}
Cardiovascular diseases (CVDs) remain the leading cause of death worldwide \cite{20252167}. Echocardiography is the clinical standard for cardiac assessment due to its non-invasive and real-time nature \cite{gillam2024echo}, making accurate left ventricle (LV) segmentation essential for hemodynamic analysis  and cardiac function quantification \cite{lang2015recommendations,Ouyang_2020_EchoNet}. However, manual contouring remains time-consuming, highly expert-dependent, and prone to significant inter-observer variability \cite{tromp2021automated,chen2020deep}.

Early studies primarily focused on 2D echocardiography segmentation, utilizing attention or pyramid modules \cite{chen2023aaunet,liu2021pyramid,guo2021dual,awasthi2022lvnet,zhou2023dsanet}. However, echocardiogram quality is often degraded by inherent speckle noise, blurred anatomical boundaries, and significant inter-individual structural variations, leading to ambiguous spatial semantic information \cite{hassan2012carotid,leclerc2019camus}. Consequently, networks need to leverage temporal information to obtain more accurate and robust semantic features. Existing methods for modeling spatiotemporal information can generally be divided into two categorizes. The first category performs temporal modeling over entire long videos, either by employing 3D networks or implementing 3D-SVD to extract low-rank motion information \cite{maani2024simlvseg,li2025semi}. However, these global-context-dependent approaches typically demand high computational and memory overhead, resulting in limited flexibility for variable-length or short clips, and reduced adapbility to streaming scenarios. The second category focuses on establishing temporal correlations for spatial features extracted by 2D networks: some studies enhance features through feature matching between adjacent frames or by using memory for temporal information propagation \cite{deng2024memsam,deng2025ncmnet,wu2022semisupervised}. Since they rely on simple frame stacking and ignore cardiac motion, semantic errors caused by speckle noise and artifacts accumulate over time, and segmentation precision remains limited by video quality. Meanwhile, lacking intermediate frame annotations makes high-performance semi-supervised spatiotemporal modeling a significant challenge.

To address these challenges, we designed and constructed a backbone network inspired by the OverLoCK concept \cite{lou2025overlock}. Spatial features are first extracted via basic convolutions to capture local details and preserve inductive bias, while spatial global dependency is achieved through the Overview-Focus mechanism during deep semantic extraction. To build robust spatiotemporal features, we noted the high correlation between local pixel displacement and semantic information. Inspired by the idea that inter-frame affinity matrices can be viewed as pixel-level transition probability distributions to implicitly capture spatiotemporal evolution \cite{wang2019learning}, we incorporate the Local Self-Similarity (LSS) perspective—that spatial layout is more stable than texture—to develop a local transition probability measure\cite{shechtman2007matching}. This helps the model better understand semantics and addresses the lack of temporal awareness in 2D networks. Our main contributions are as follows:
\begin{enumerate}
    \item We define a Spatio-temporal Local Self-Similarity Fusion (STLSF) module to guide semantic-texture fusion. This avoids sole reliance on texture, enabling stable correction of semantic errors through spatiotemporal transition similarities.
    \item We propose the Consistency-Informed Semantic-Refinement Network (CISR-Net), a semi-supervised framework that achieves global spatio-temporal dependency by integrating three-stage spatial feature extraction with STLSF module, ensuring superior segmentation consistency.
    \item We design a Frequency-aware Denoising pre-training (FD) strategy that integrates ultrasound-inherent imaging patterns as priors into a convolution-based, overview-focused encoder, enhancing anatomical feature extraction from unlabeled data and improving the network's robustness against noise.
\end{enumerate}

\begin{figure}[htbp]
    \centering
    \includegraphics[width=\textwidth]{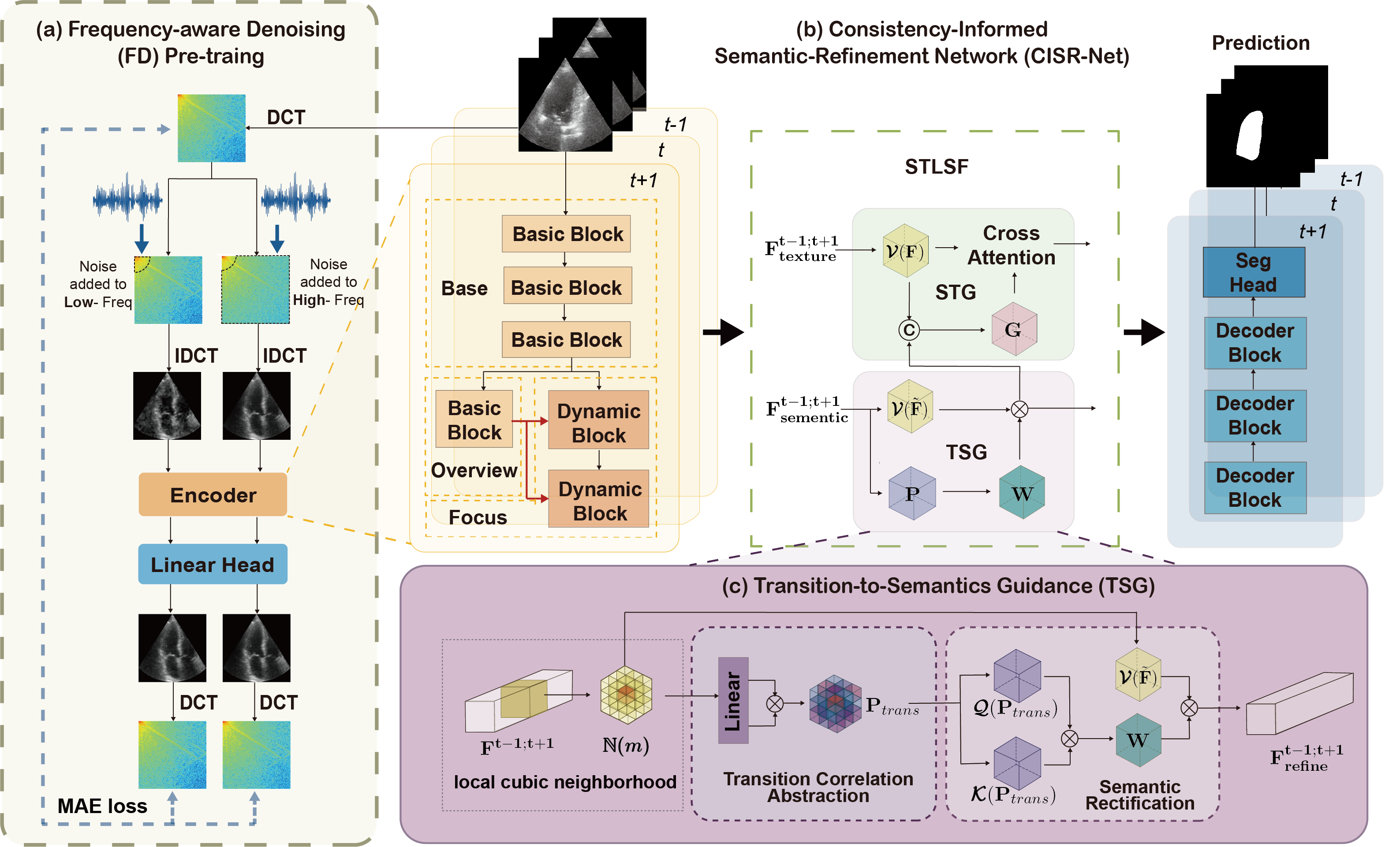}%
    \caption{Overall architecture of the FD pre-training framework and CISR-Net.}
    \label{fig:methodology_overview}
\end{figure}

\section{Methodology}
\subsection{Overview}
The proposed framework is illustrated in Fig.~\ref{fig:methodology_overview}. First, the temporal dimension of input sequences is collapsed into the batch dimension for multi-scale spatial feature extraction via a 2D encoder. These hierarchical features are then processed by our core STLSF module to establish robust spatiotemporal dependencies and rectify noise-induced semantic errors. Finally, a convolutional decoder aggregates the enhanced features for pixel-wise prediction. Additionally, a FD pre-training strategy (Fig.~\ref{fig:methodology_overview}a) initializes the encoder, tailoring the backbone to ultrasound-specific characteristics.

\subsection{Frequency-aware Denoising Pre-training}

To improve robustness against complex speckle noise, we explicitly integrate ultrasound-inherent imaging patterns as priors into the convolution-based encoder $\mathcal{E}$ by introducing frequency-domain perturbations. The corrupted input $\hat{\mathbf{X}}$ and the frequency-aware denoising objective $\mathcal{L}$ are defined as:
\begin{align}
    \hat{\mathbf{X}} &= \text{IDCT} \left( X_{\text{freq}} + \lambda \cdot (N \odot M) \right), \\
    \mathcal{L} &=  \left|\left( \text{DCT}\left( \text{Linear}(\mathcal{E}(\hat{\mathbf{X}})) \right) - X_{\text{freq}} \right) \odot M \right|
\end{align}
where $X_{\text{freq}}, N, M, |\cdot|,$ and $\hat{\mathbf{X}}$ denote the clean image Discrete Cosine Transform (DCT) spectrum, the $1/f$ ultrasound speckle noise map, the frequency band mask, the MAE loss, and the noisy image reconstructed by the inverse DCT (IDCT) from the modified frequency spectrum, respectively. $\lambda$ scales noise intensity, and $\text{Linear}(\cdot)$ is the projection head.

To enhance global dependencies, the encoder $\mathcal{E}$ employs a \textit{base-overview-focus} framework. Initially, the \textbf{base} and \textbf{overview} phases sequentially extract shallow local features and coarse deep semantics via Basic Blocks $\mathcal{B}(\cdot)$:
\begin{equation}
    \mathbf{F}_i = \mathcal{B}(\mathbf{F}_{i-1}), \quad i \in \{1, 2, 3, 4\}
\end{equation}
where $\mathbf{F}_0 = \hat{\mathbf{X}}$. Here, $\{\mathbf{F}_1, \mathbf{F}_2,\mathbf{F}_3\}$ represent the shallow local details extracted in the base phase, while $\{\mathbf{F}_4\}$ provide a semantic overview. In the \textbf{focus} phase, Dynamic Blocks $\mathcal{D}(\cdot)$ utilize the overview $\mathbf{F}_4$ as a structural prompt for precise global extraction and deep feature aggregation:
\begin{align}
    \tilde{\mathbf{F}}_3 &= \mathcal{D}(\mathbf{F}_3, \mathbf{F}_4), \\
    \tilde{\mathbf{F}}_4 &= \mathcal{D}(\mathbf{F}_4, \tilde{\mathbf{F}}_3)
\end{align}
This interactive fusion refines the comprehensive representation, outputting the final encoded feature as $\mathcal{E}(\hat{\mathbf{X}}) = \tilde{\mathbf{F}}_4$. These hierarchical representations are then forwarded to the STLSF module, outputting the multi-scale encoded features as $\{\mathbf{F}_1, \mathbf{F}_2, \tilde{\mathbf{F}}_3, \tilde{\mathbf{F}}_4\}$.

\subsection{Spatio-temporal Local Self-Similarity Fusion}
This module operates in a coarse-to-fine manner: it first utilizes local transition probability correlation in the deep layers to stabilize and rectify semantic representations, and subsequently uses these refined semantics to filter noise in the shallow texture layers. 

Since pixel motions typically occur within a local region, our STLSF module confines spatio-temporal interactions to a local neighborhood. Given a 3D spatio-temporal feature map of size $x \times y \times t$, for any query point $m$, we define a point set $\mathbb{N}(m)$ to represent the neighboring points involved in the computation, which consists of all points within a local window of size $k$ centered at $m$.

\subsubsection{Transition-to-Semantics Guidance}

As illustrated in Fig.~\ref{fig:methodology_overview}(c), deep semantic features often suffer from temporal discontinuity due to speckle noise. The TSG module reduces this semantic ambiguity by leveraging stable spatiotemporal motion patterns through two phases.

\paragraph{Transition Correlation Abstraction Phase}
For any query point $m$ within the deep semantic features $\tilde{\mathbf{F}}$, we define a local transition probability distribution $\mathbf{P}_{trans}(m)$ to implicitly represent cardiac motion. This distribution is derived from the transition affinity between $m$ and its neighborhood $n \in \mathbb{N}(m)$, computed via independent linear projections and normalized by a softmax function:
\begin{equation}
    \mathbf{P}_{trans}(m, n) = \text{Softmax}_{n \in \mathbb{N}(m)}\left( \text{Linear}_1(\tilde{\mathbf{F}}_m) \cdot \text{Linear}_2(\tilde{\mathbf{F}}_n)^\top \right)
\end{equation}
where $\text{Linear}_1(\cdot)$ and $\text{Linear}_2(\cdot)$ are distinct projection layers.

\paragraph{Semantic Rectification Phase}
We then utilize the structural consistency of these transition probabilities to guide feature rectification via an attention formulation:
\begin{equation}
    \tilde{\mathbf{F}}_{final}(m) = \text{Attn} \left( \mathcal{Q}(\mathbf{P}_{trans}(m)), \mathcal{K}(\mathbf{P}_{trans}(n)), \mathcal{V}(\tilde{\mathbf{F}}_n) \right)_{n \in \mathbb{N}(m)}
\end{equation}
In this formulation, the internal attention affinity inherently serves as a structural descriptor, evaluating the similarity between the local transition probability distributions $\mathbf{P}_{trans}$. While the raw encoded feature $\tilde{\mathbf{F}}$ is susceptible to imaging artifacts, this transition-guided attention aligns semantics based on structural coherence, effectively rectifying inconsistencies caused by localized noise.

\subsubsection{Semantics-to-Texture Guidance}
Shallow ultrasound features $\mathbf{F}_i$ are highly susceptible to speckle noise. To enhance boundary details while preserving semantic consistency, the STG module fuses the noisy shallow features $\mathbf{F}_i$ with stable deep features $\mathbf{D}_{i+1}$ (where $i \in \{1, 2\}$ is the resolution level). We first generate a semantic-guided multi-level feature $\mathbf{G}$:
\begin{equation}
    \mathbf{G} = \mathcal{F}_{\text{proj}}\left( \text{Concat}\left( \mathbf{F}_i, \text{Up}(\mathbf{D}_{i+1}) \right) \right)
\end{equation}
where $\mathcal{F}_{\text{proj}}$ is a standard projection layer. To achieve a large receptive field efficiently, the refined feature $\hat{\mathbf{F}}_i$ is computed via a local cross-attention mechanism within a dilated spatiotemporal cubic neighborhood $\mathbb{N}_d(m)$:
\begin{equation}
    \hat{\mathbf{F}}_i(m) = \text{CrossAttn}\left( \mathcal{Q}(\mathbf{G}_m), \mathcal{K}(\mathbf{G}_n), \mathcal{V}(\mathbf{F}_{i,n}) \right)_{n \in \mathbb{N}_d(m)}
\end{equation}
Here, $\text{CrossAttn}(\cdot)$ denotes the standard scaled dot-product attention. By leveraging the fused semantic prior $\mathbf{G}$ to query and weight the local textural values of $\mathbf{F}_i$, this module yields boundary-consistent and anatomically precise segmentation masks.

\section{Experiments and Results}
\begin{table}[t]
    \centering
    \caption{Quantitative comparison with state-of-the-art methods on CAMUS and EchoNet-Dynamic datasets.}
    \label{tab:quant_results}
    \small
    \begin{tabular}{lccccccc}
        \toprule
        \multirow{2}{*}{\textbf{Method}} & \multirow{2}{*}{\textbf{Venue/Year}} & \multicolumn{3}{c}{\textbf{CAMUS}} & \multicolumn{3}{c}{\textbf{EchoNet-Dynamic}} \\
        \cmidrule(lr){3-5} \cmidrule(lr){6-8} 
        & & Dice $\uparrow$ & HD95 $\downarrow$ & ASSD $\downarrow$ & Dice $\uparrow$ & HD95 $\downarrow$ & ASSD $\downarrow$ \\
        \midrule
        XMem++~\cite{bekuzarov2023xmem}    & ICCV 2023 & 93.45 & 3.45 & 1.42 & 87.72 & 3.64 & 1.72  \\
        VideoMamba~\cite{li2024videomamba} & ECCV 2024 & 91.26 & 6.28 & 2.67 & 90.01 & 4.59 & 1.77 \\
        H2Former~\cite{he2023h2former}     & TMI 2023   & 92.34 & 4.75 & 1.86 & 91.89 & 4.67 & 1.57 \\
        EchoONE~\cite{hu2025echoone}       & CVPR 2025 & 93.07 & 3.85 & 1.61 & 92.12 & 5.89 & 2.48 \\
        PKEcho-Net~\cite{wu2023pkecho}     & AAAI 2023 & 93.23 & 6.03 & 2.46 & 91.89 & 3.61 & 1.60  \\
        MemSAM~\cite{deng2024memsam}       & CVPR 2024 & 92.88 & 4.09 & 1.65 & 92.26 & 6.40 & 2.59 \\
        SimLVSeg~\cite{maani2024simlvseg}  & UMB 2024   & 92.88 & 4.69 & 1.75 & 92.11 & 3.91 & 1.40\\
        NCMNet~\cite{deng2025ncmnet}        & TMI 2025  & 93.56 & 4.08 & 1.83 & 92.30 & 4.04 & 1.47 \\
        \midrule
        \textbf{Ours} & \textbf{-} & \textbf{93.87} & \textbf{3.29} & \textbf{1.33} & \textbf{92.62} & \textbf{2.73} & \textbf{1.11} \\
        \bottomrule
    \end{tabular}
\end{table}
\subsection{Datasets and Pre-processing}
To ensure a fair comparison, we uniformly evaluate all models on two public benchmarks: CAMUS~\cite{leclerc2019camus} and EchoNet-Dynamic~\cite{Ouyang_2020_EchoNet}. For data partitioning, we adopt a 7:1:2 train/val/test split for CAMUS and strictly follow the official split for EchoNet-Dynamic. During segmentation training, video clips of $T=10$ frames are sampled, where only the End-Diastolic (ED) and End-Systolic (ES) frames provide pixel-level annotations. For self-supervised pre-training, 5 frames are randomly sampled from each video exclusively from the training set to prevent data leakage. All frames are resized to $224 \times 224$ and $128 \times 128$ for CAMUS and EchoNet-Dynamic, respectively. Standard online augmentations, including random rotation, horizontal flipping, and brightness adjustment, are applied to enhance model robustness.

\subsection{Implementation Details}
Our strategy consists of two stages: FD pre-training and downstream fine-tuning. The encoder is first pre-trained for 500 epochs with batch size 32, learning rate $1 \times 10^{-3}$. During fine-tuning, the encoder is initialized with pre-trained weights and optimized via AdamW for 100 epochs. We employ a cosine annealing scheduler with initial learning rate $5 \times 10^{-3}$, weight decay 0.05 and a 10-epoch linear warm-up. Following the semi-supervised protocol, the joint loss $\mathcal{L}_{total} = 0.8\mathcal{L}_{bce} + 1.2\mathcal{L}_{dice}$ is computed only on labeled ED and ES frames, while intermediate frames provide temporal constraints. The optimal model is selected based on the peak validation Dice score. All experiments are conducted in PyTorch on a single NVIDIA RTX 4090 GPU.Our model achieves 78.62 G FLOPs and 38.5 MB parameters, with a processing speed of 15.07 clips/s (150 frames/s) and a latency of 66.35 ms/clip.

\subsection{Evaluation Metrics}
To quantitatively evaluate segmentation performance, we adopt three standard metrics: Dice, HD95, and ASSD. Dice measures the overlap between predictions and ground truth, while HD95 and ASSD assess boundary accuracy and shape consistency (in mm). Specifically, HD95 reports the 95th percentile boundary distance to reduce the influence of outliers, and ASSD computes the average symmetric surface distance. During testing, all annotated frames are evaluated on CAMUS, whereas only ED and ES frames are used for EchoNet-Dynamic due to its labeling protocol. Higher Dice and lower HD95/ASSD indicate better performance.

\subsection{Comparison with State-of-the-art Methods}
We benchmark our framework against eight baselines, ranging from generic video models (XMem++~\cite{bekuzarov2023xmem}, VideoMamba~\cite{li2024videomamba}) to medical-specific networks (H2Former~\cite{he2023h2former}, EchoONE~\cite{hu2025echoone}), and specialized video approaches (MemSAM~\cite{deng2024memsam}, NCMNet~\cite{deng2025ncmnet}, PKEcho-Net~\cite{wu2023pkecho}, SimLVSeg~\cite{maani2024simlvseg}).As summarized in Table~\ref{tab:quant_results}, our method consistently achieves SOTA performance across two public datasets.

\begin{figure}[t]
    \centering
    \includegraphics[width=0.9\textwidth]{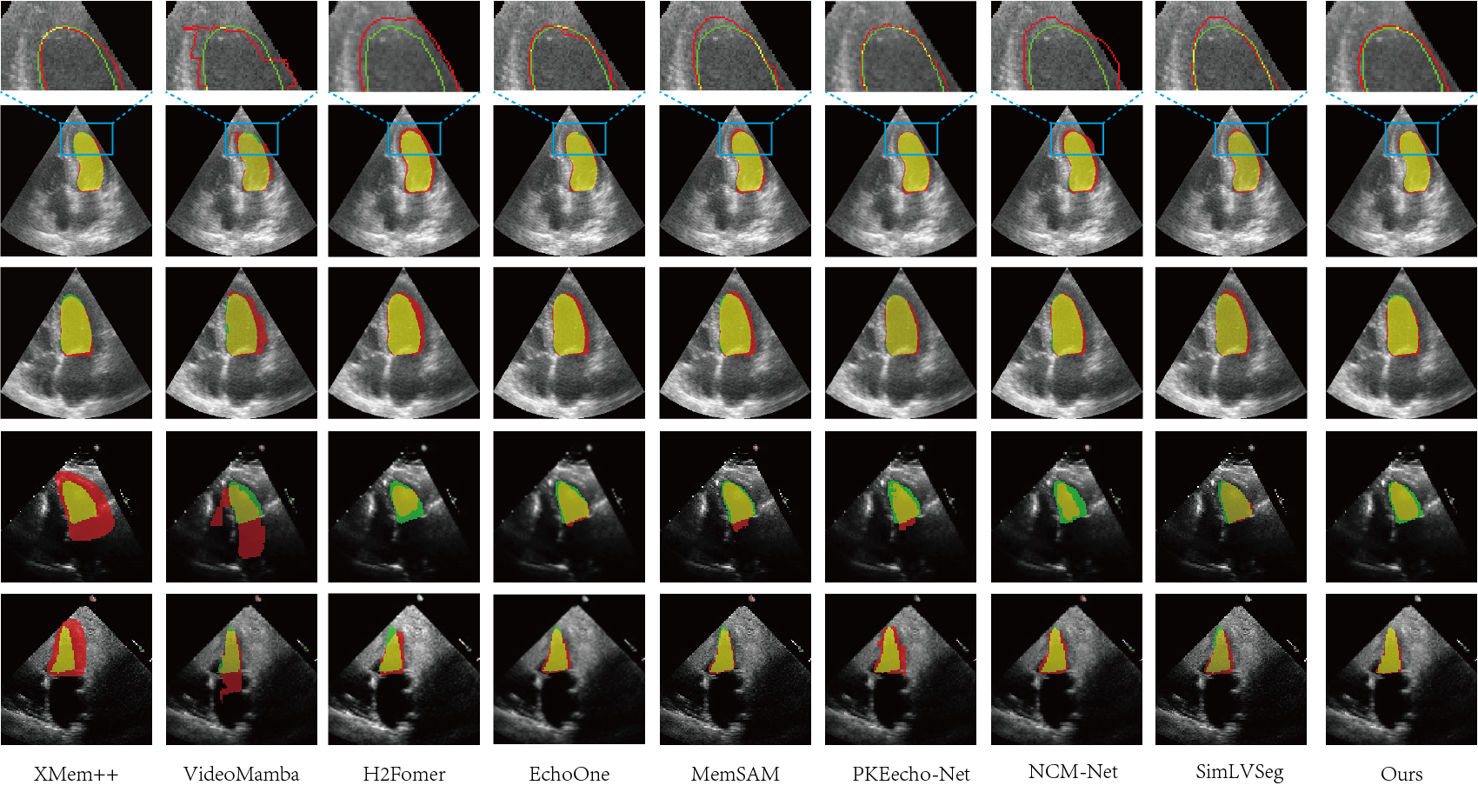}
    \caption{Visual comparison with representative baselines on CAMUS and EchoNet-Dynamic datasets.}
    \label{fig:visual}
\end{figure}

As shown in Fig.~\ref{fig:visual}, we present two representative cases from the CAMUS and EchoNet-Dynamic datasets, respectively. In the visualization, predictions are shown in red, ground-truth masks in green, and their overlapping regions in yellow. Owing to the lack of domain-specific modules tailored for echocardiography, generic models struggle to accurately extract cardiac structures. The predictions of single-frame models exhibit noticeable aliasing artifacts (serrated boundaries). Other specialized models also suffer from limited temporal continuity and substantial semantic errors.
In contrast, benefiting from the incorporation of the STLSF module, our model produces smoother segmentation boundaries. Furthermore, with the noise adaptation capability and anatomical cardiac understanding acquired during pre-training, our framework maintains precise segmentation performance even in challenging low-quality cases from the EchoNet-Dynamic dataset.

\subsection{Ablation Study}

\subsubsection{Contribution of STLSF Modules.} Table~\ref{tab:module_ablation}(a) investigates the STLSF module. Compared to a baseline passing unrefined features to the decoder, integrating TSG and STG individually improves Dice score by 0.26\% and 0.24\%. This validates our dual-path design: utilizing local transition probability correlations to rectify semantics, followed by semantic-guided filtering for textural refinement, which together yield a robust architecture for spatiotemporal coherence.
\subsubsection{Effectiveness of Learning Strategies.} 
As shown in Table~\ref{tab:module_ablation}(b), we compared the FD pre-training strategy with several auxiliary tasks based on data augmentation, including super-resolution (SR), deblurring\cite{chen2021pre}, denoising\cite{vincent2008extracting}, and masked frequency modeling (MFM)\cite{xie2023masked}, based on average results across two public datasets. Among these tasks, Denoising and MFM achieved good results (93.02\%, 93.03\%), showing that learning noise adaptation and frequency domain features is indeed beneficial. Our FD strategy achieved the best result (93.19\%). By shifting noise injection from the spatial domain to the frequency domain, the model is guided to adapt to the spectral distribution, thereby learning an encoder with noise robustness.
\begin{table}[t]
    \centering
    \caption{Ablation studies averaged over CAMUS and EchoNet-Dynamic datasets.}
    \label{tab:module_ablation}
    \small 
    \begin{subtable}[b]{0.48\textwidth}
        \centering
        \caption{STLSF components}
        \setlength{\tabcolsep}{4pt}
        \begin{tabular}{l c c}
            \toprule
            Method & Dice (\%) & HD95(mm) \\
            \midrule
            Baseline & 92.75 & 4.05 \\
            + TSG & 93.01 (+0.26) & 3.68 (-0.37) \\
            + STG & 92.99 (+0.24) & 3.72 (-0.33) \\
            \midrule
            \textbf{Ours} & \textbf{93.25 (+0.50)} & \textbf{3.01 (-1.04)} \\
            \bottomrule
        \end{tabular}
    \end{subtable}
    \hfill 
    \begin{subtable}[b]{0.48\textwidth}
        \centering
        \caption{Pre-training strategies}
        \setlength{\tabcolsep}{4pt}
        \begin{tabular}{l c c}
            \toprule
            Strategy & Dice (\%) & HD95(mm) \\
            \midrule
            w/o SR & 92.87 & 3.76 \\
            w/o Deblur & 92.92 & 3.78 \\
            w/o Denoise & 93.02 & 3.67 \\
            w/o MFM & 93.03 & 3.65 \\
            \midrule
            \textbf{Ours} & \textbf{93.19} & \textbf{3.01} \\
            \bottomrule
        \end{tabular}
    \end{subtable}
\end{table}
\section{Conclusion}
In this paper, we propose a semi-supervised framework for echocardiogram video segmentation to improve robustness against noise and ambiguous semantics via stable pixel-level similarity transfer. An efficient CNN with local inductive bias is adopted as the backbone. Combined with a three-stage overview-focus architecture and the STLSF module, the framework captures global spatio-temporal dependencies while maintaining low computational cost and temporal consistency. In addition, the FD pre-training strategy enhances noise robustness by adapting the encoder to the spectral characteristics of ultrasound data.
Extensive experiments on CAMUS and EchoNet-Dynamic demonstrate state-of-the-art performance in terms of Dice and boundary metrics (HD95/ASSD). Experimental results suggest that the proposed framework provides improved stability in scenarios involving rapid cardiac motion and severe speckle noise. By alleviating the limitations of scarce annotations, this work highlights the potential of leveraging spatio-temporal cardiac dynamics for more reliable automated cardiac assessment.

\begin{credits}
\subsubsection{\ackname}
This research was partly supported by the National Natural Science Foundation of China (Grant No. 62501142) and the Natural Science Foundation of Jiangsu Province (Grant No. BK20241277).

\subsubsection{\discintname}
The authors have no competing interests to declare that are relevant to the content of this article.
\end{credits}

\bibliographystyle{splncs04} 
\bibliography{refs}   

\end{document}